\documentclass[conference]{IEEEtran}

\usepackage{expl3}
\ExplSyntaxOn
\newcommand\latinabbrev[1]{
	\peek_meaning:NTF . {
		#1\@}%
	{ \peek_catcode:NTF a {
			#1.\@ }%
		{#1.\@}}}
\ExplSyntaxOff

\def\etal{\latinabbrev{et al}}

\def\ie{\latinabbrev{i.e}}
\usepackage{graphicx}
\usepackage{algorithm}
\usepackage{algorithmic}

\usepackage{amssymb}
\usepackage{letltxmacro}
\usepackage{pifont}
\usepackage[cmex10]{amsmath}
\usepackage{algorithmic}
\usepackage{color}
\usepackage{cases}
\usepackage{bm}
\usepackage[caption = false]{subfig}
\usepackage{multirow}

\usepackage{lipsum}
\usepackage{tikz}
\def\checkmark{\tikz\fill[scale=0.4](0,.35) -- (.25,0) -- (1,.7) -- (.25,.15) -- cycle;}

\ifCLASSINFOpdf
\else
\fi
\begin{document}
%
\title{End-to-End Adversarial Learning for \\ 
	Intrusion Detection in Computer Networks}


\author{Bahram~Mohammadi$^1$,
	and~Mohammad~Sabokrou$^2$
	\\ $^1$Sharif University of Technology~~ $^2$Institute for Research in Fundamental Sciences (IPM)}

\maketitle

\begin{abstract}
This paper presents a simple yet efficient method for an anomaly-based Intrusion Detection System (IDS). In reality, IDSs can be defined as a one-class classification system, where the normal traffic is the target class. The high diversity of network attacks in addition to the need for generalization, motivate us to propose a semi-supervised method. Inspired by the successes of Generative Adversarial Networks (GANs) for training deep models in semi-unsupervised setting, we have proposed an end-to-end deep architecture for IDS. The proposed architecture is composed of two deep networks, each of which trained by competing with each other to understand the underlying concept of the normal traffic class. The key idea of this paper is to compensate the lack of anomalous traffic by approximately obtain them from normal flows. In this case, our method is not biased towards the available intrusions in the training set leading to more accurate detection. The proposed method has been evaluated on NSL-KDD dataset. The results confirm that our method outperforms the other state-of-the-art approaches. 

\end{abstract}


%
\IEEEpeerreviewmaketitle

\section{Introduction}
The significant Internet expansion and also its rising popularity cause a massive increase in data exchange between different parties. These data most probably include valuable information of people, government, and organizations. Therefore, a reliable defense system is required 
to prevent misuse of sensitive information and to detect network vulnerabilities and potential threats. 
Intrusion Detection System (IDS) is a promising solution for providing network security services. Also, it is considered an effective alternative for conventional defense systems such as firewall. Previous researches in this topic, can be divided into two major categories: (1) signature-based, and (2) anomaly-based \cite{signature_anomaly}. Signature-based approach is very effective to detect already known attacks which their patterns are available. However, they are inefficient against unfamiliar and new intrusions.
Conversely, the latter approach seems more effective owing to its superior performance against unknown and zero-day network attacks \cite{anomaly_based}. 

Generally, supervised methods need to know the specific characteristics of attacks, while their high diversity makes this process expensive and probably impossible \cite{supervised_diversity}. This difficulty makes supervised manner lacks generalization. To overcome such problem, we have proposed a semi-supervised method. In contrary to supervised methods which detection is performed based on the available labeled training data, our method is able to decide about the unseen incoming network Packet Flows (PFs). 


In recent years, Deep Neural Networks (DNNs) have achieved state-of-the-art performance in various research fields, especially in computer vision and natural language processing. These outstanding successes motivate us to take advantages of deep learning in IDS. Note that, DNNs such as Convolutional Neural Networks (CNNs), can achieve the remarkable results only if they access to a lot of annotated training samples from all classes. Under the realistic conditions, there are numerous samples from normal class, but the abnormal class is often absent during the training, poorly sampled, or not well defined. In summary, the anomaly detection task refers to a binary classification task, while there are not any samples from abnormal/outlier class. Training an end-to-end DNN in the absence of one class data, is not straightforward \cite{sabokrou2018adversarially}.

To address the above-mentioned challenge, We have aimed to generate simulated anomalous flows, inspired by the Generative Adversarial Networks (GANs) \cite{gan}. Using GAN, we have proposed an end-to-end deep network for IDS which is able to effectively detect the network intrusions, even against unforeseen and new ones.  
The proposed method is composed of two main modules, $\mathcal{R}$econstructor network ($\mathcal{R}$), and $\mathcal{A}$nomaly detector network ($\mathcal{A}$). Since $\mathcal{R}$ and $\mathcal{A}$ are trained adversarially and competitively, they properly learn the distribution of the feature space of normal PFs. $\mathcal{R}$ strives to fool $\mathcal{A}$, in efforts to manipulate it into detecting $\mathcal{R}(X)$ as a normal PF not a reconstructed version, however, the training duration of $\mathcal{R}$ network is limited in order to have more realistic anomalies. On the other hand, The duty of $\mathcal{A}$ is to distinguish between original normal PFs and the reconstructed ones (abnormal traffic). In other words, $\mathcal{A}$ determines whether the incoming PF, follows the distribution of the feature space of the normal traffic. 

The main contributions of the proposed method are: (1) Proposing an end-to-end DNN for IDS which is trained adversarially in a GAN-style setting. To the best of our knowledge, our method is the first end-to-end DNN for semi-supervised intrusion detection, (2) In our method, training is done merely using the feature space of the normal PFs. Nevertheless, the experimental results show the superiority of the proposed method, even compared to the supervised approaches, due to its generality. (3) Performance of the proposed method is better than the other state-of-the-arts in terms of accuracy while the time complexity is also reduced noticeably. (4) In the proposed method, the unseen traffic, \ie,~ anomaly, is simulationly generated using adversarial training. Hence, the need for availability of all classes data has been satisfied.
\begin{figure}
	\subfloat[Training Phase]{
		\includegraphics[clip,width=\columnwidth]{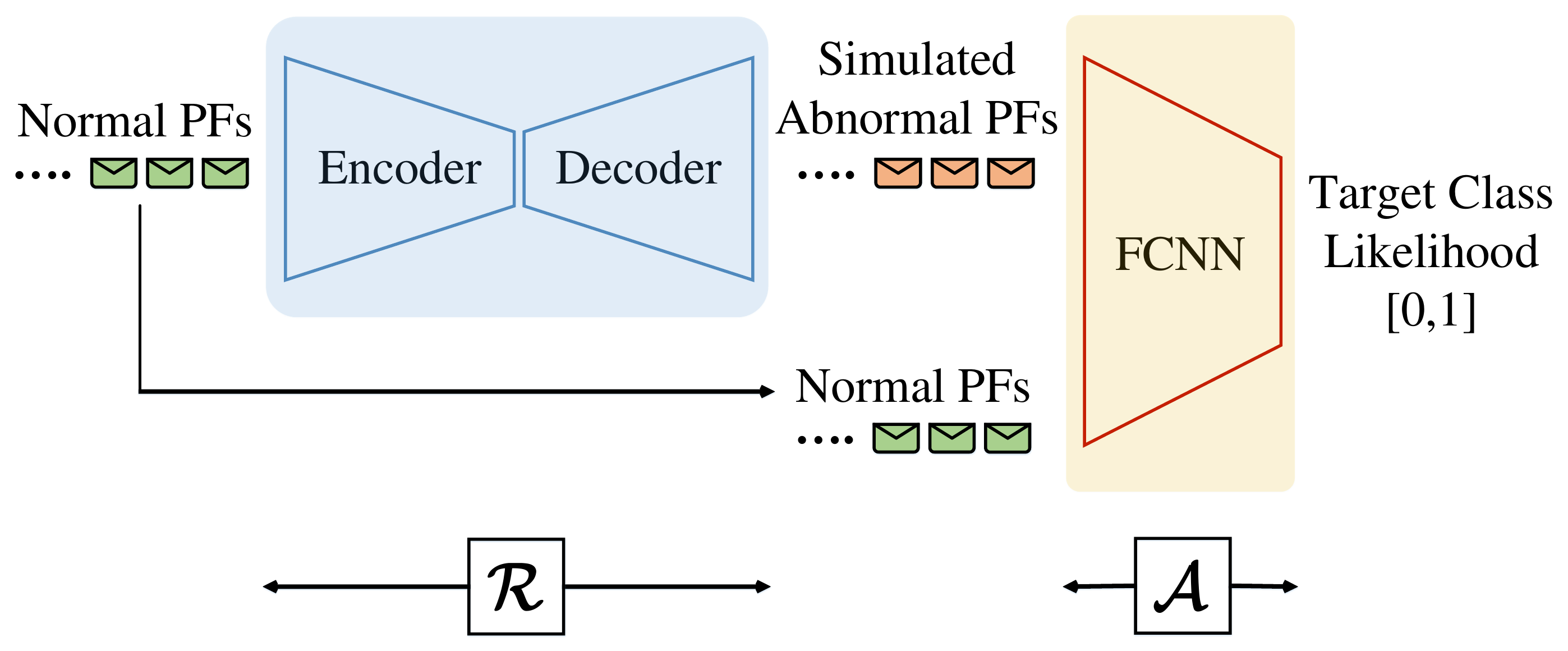}
	}
	
	\subfloat[Test Phase]{
		\includegraphics[clip,width=\columnwidth]{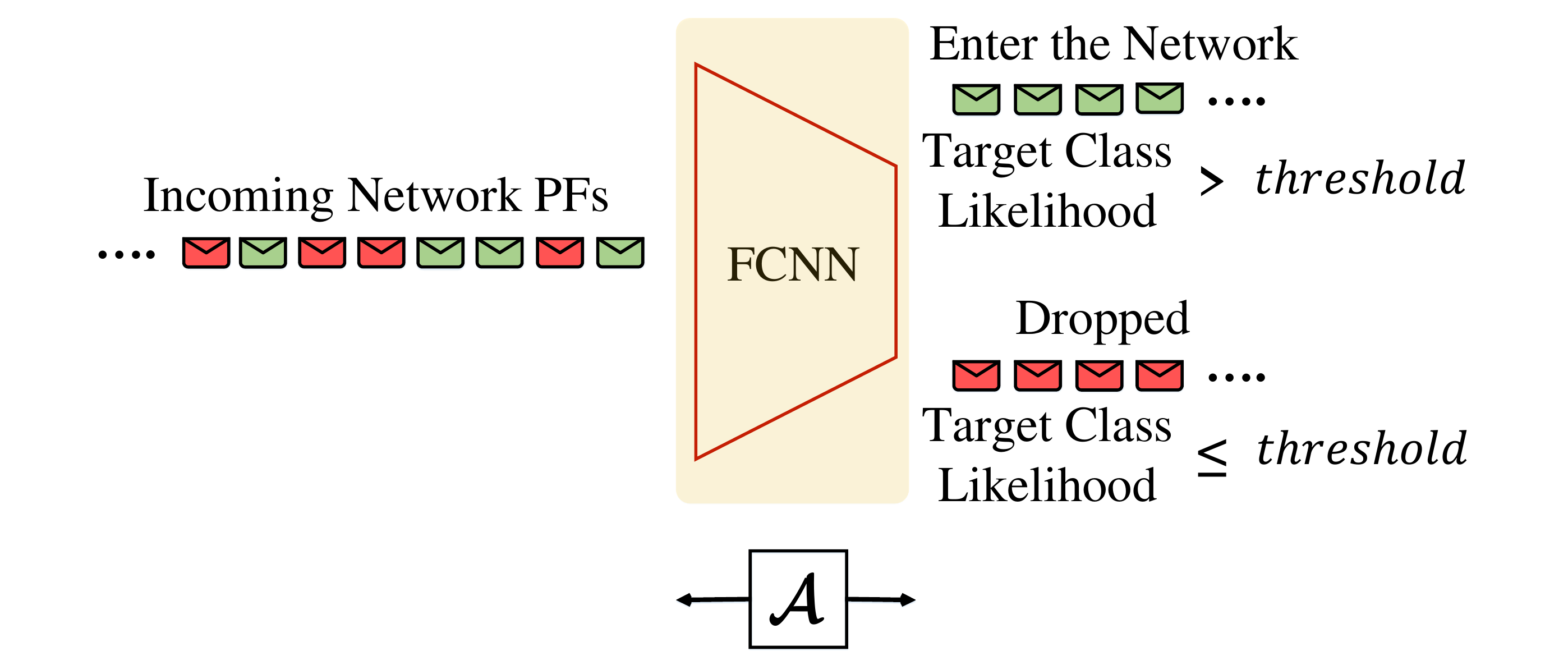}
	}
	\caption{Overall scheme of the proposed method for anomaly detection in computer networks in the stages of training and test. $\mathcal{R}$ and $\mathcal{A}$ are two modules of the model which are trained adversarially and competitively. $\mathcal{R}$ includes an encoder-decoder network and $\mathcal{A}$ consists of a Fully-Connected Neural Network (FCNN) ending with a softmax classifier. In training phase, $\mathcal{R}$ parameters are optimized to reconstruct the incoming normal PFs for generating simulated abnormal traffic while it attempts to achieve an optimum value for the reconstruction error. Additionally, $\mathcal{A}$ can detect if an incoming network PF belongs to the target class (normal) or it is an outlier (anomaly). In test phase, The classification is performed based on a predefined $threshold$. The simulated abnormal PFs have some deviation from the real anomalies. Thus, this fact is pointed out by using different colors for simulated abnormal  and real abnormal packets. 
	}
	\label{fig:proposed_method}
\end{figure}

\section{End-to-end IDS}
\label{sec:end-to-end}
In this paper, we have proposed a semi-supervised and end-to-end deep model for IDS. Accordingly, the abnormal class data is not available. The key idea of the proposed method is to solve this problem by simulating anomalous PFs using original normal traffic. This method comprises two modules (\ie,~ networks): (1) $\mathcal{R}$,  and (2) $\mathcal{A}$. The former generates simulated anomalous PFs by reconstructing normal traffic to obviate the need for the presence of anomaly class in training phase. The latter detects whether the incoming traffic is normal or not. $\mathcal{R}$ and $\mathcal{A}$ are trained adversarially and unsupervisedly in an end-to-end setting. $\mathcal{R}$ reconstructs the incoming PFs to mislead $\mathcal{A}$ over the input type, \ie,~ normal flow or simulated abnormal one. Since the original data is available for $\mathcal{A}$, it is familiar with their concept. Therefore, $\mathcal{A}$ does not act blindly and attempts to reject the simulated anomalous traffic. These two modules are trained in a GAN-style setting, forming an end-to-end framework for anomaly detection in IDS. The difference of our work with the original GAN is the time which the generator training process stops. $\mathcal{R}$ should not perfectly reconstruct the normal traffic. In our case, simulated anomalies should have some deviation from the normal traffic, otherwise, anomalous PFs are not properly generated. $\mathcal{A}$ can also effectively distinguish between real and fake input. After training process, $\mathcal{A}$ knows the distribution of the target class, \ie,~ normal traffic, and can simply investigate that each of new incoming flow follows the distribution or not. Fig. \ref{fig:proposed_method} shows an overview of our proposed method. A detailed explanation  of $\mathcal{R}$ network, $\mathcal{A}$ network, joint training of $\mathcal{R}+\mathcal{A}$, and the process of anomaly detection are also provided in this section.

\textbf{$\mathcal{R}$econstructor Network:}
\label{sec:R_network}
$\mathcal{R}$ has simulationly and adverserially generated anomalous traffic in the stage of training by reconstructing normal incoming PFs. This network gradually and in competition with $\mathcal{A}$, learns to generate flows similar to normal ones. To this end, $\mathcal{R}$ includes an encoder-decoder network. Equation \ref{eq:autoencoder_encoder_decoder} shows the function of these components.
\begin{equation}
	\begin{split}
		\label{eq:autoencoder_encoder_decoder}
		\text{}\begin{cases}
			\text{Encoder:} & \text{$Z$ $=$ $\sigma(WX+b)$} \\
			\text{Decoder:} & \text{$X'$ $=$ $\sigma(W'Z+b')$}
		\end{cases}
	\end{split} 
\end{equation}

\begin{figure}
	\includegraphics[width=\linewidth]{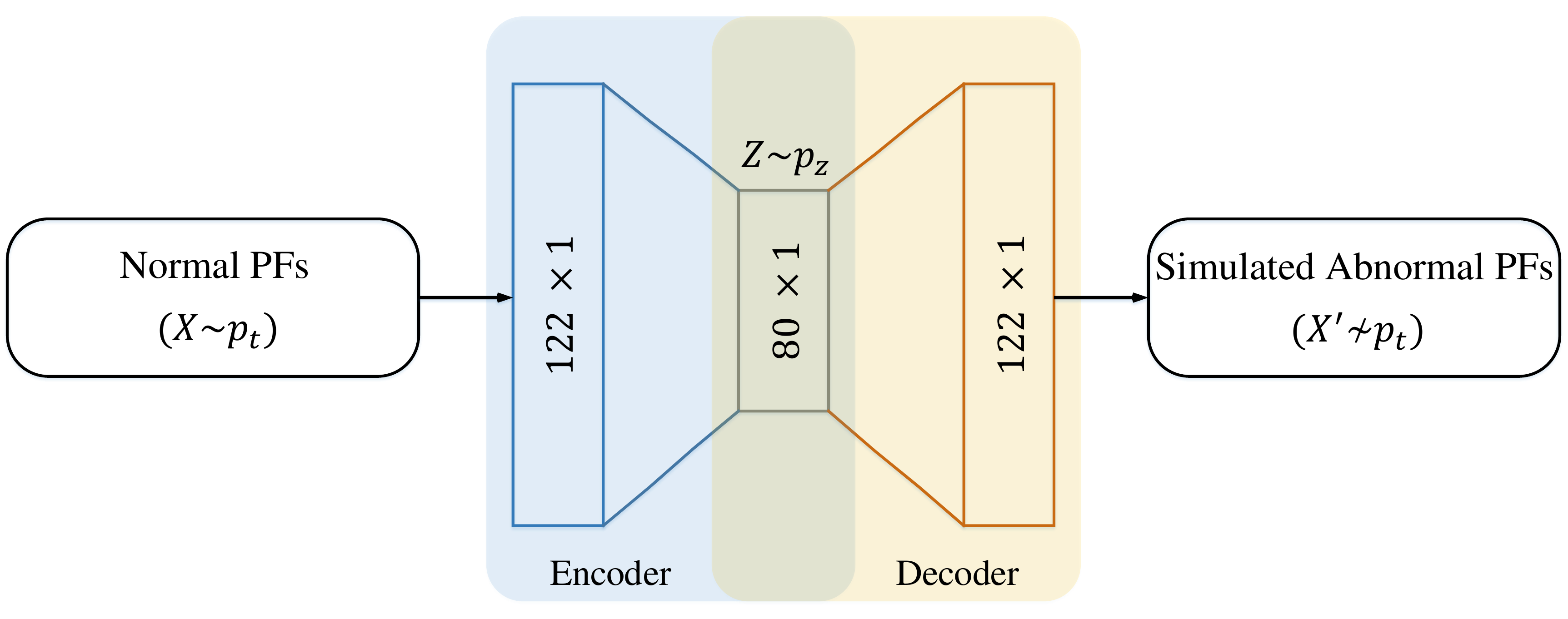}
	\vspace{-5mm}
	\caption{The architecture of $\mathcal{R}$ encoder-decoder network.}
	\label{fig:R_network}
\end{figure}

Here, $ X \in \mathbb{R}^n $ is the feature space of the incoming PF, $X' \in \mathbb{R}^n$ is the feature space of the simulated abnormal PF, $Z \in \mathbb{R}^m$ is the latent representation, $W$ and $W'$ are weight matrices, $\sigma$ and $\sigma'$ are element-wise activation functions. The encoder maps an incoming PF to a latent space and the decoder attempts to retrieve the simulated anomaly from the latent space. Note that, if $\mathcal{R}$ reconstructs the normal flows with high precision, they can not play the role of anomalies. Consequently, the training of $\mathcal{R}$ should be stopped before it be able to perfectly reconstruct the normal flows. In \cite{sabokrou2016video,sabokrou2018adversarially}, it has been shown that, by over-training an encoder-decoder network, we could inpatient the irregular samples and convert them into a normal concept. In contrary to these works, in our case, it is not desirable that $\mathcal{R}$ encoder-decoder maps the flows including irregularity to equivalent normal flows by their reconstruction. 
%
%
Fig. \ref{fig:R_network} shows the architecture of the $\mathcal{R}$ network. 

\begin{figure}
	\includegraphics[width=\linewidth]{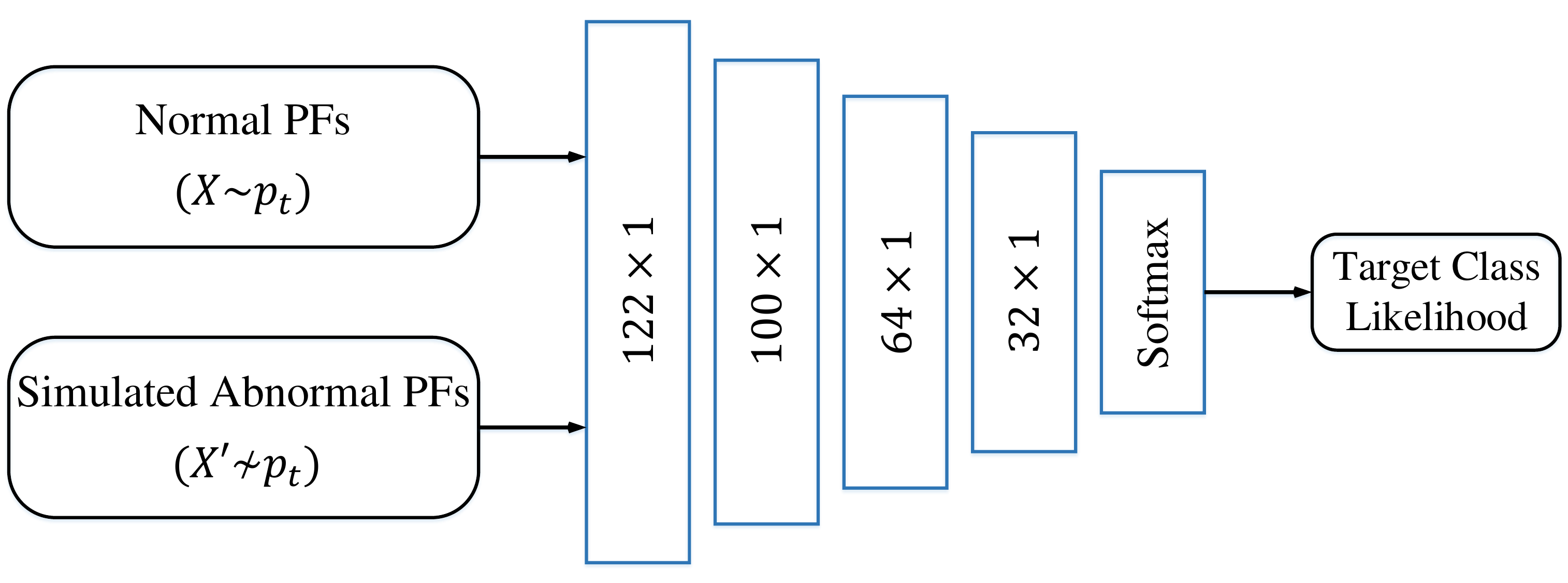}
	\vspace{-6mm}
	\caption{$\mathcal{A}$ network architecture specifying if an incoming PF follows the target class distribution or not.}
	\label{fig:A_network}
\end{figure}

\textbf{$\mathcal{A}$nomaly detector Network:}
\label{sec:A_network}
The $\mathcal{A}$ network acts as a classifier to distinguish between the representation of simulated traffic \ie,~ abnormal/fake flows, generated by $\mathcal{R}$ and normal traffic \ie,~real flows. Previously, in computer vision community \cite{sabokrou2018adversarially,sabokrouavid}, it is investigated that a network like $\mathcal{A}$, which somehow differs from ours in terms of learning process, is capable of efficiently detecting the irregular/outlier images. 
$\mathcal{A}$ includes a sequence of fully-connected layers ending with a softmax layer (classifier).  
As mentioned previously, the main purpose of $\mathcal{A}$ is to detect the abnormal incoming PFs. It is worth mentioning that, the $\mathcal{A}$ output indicates the likelihood of the input following the distribution spanned by the target class. Hence, outputs can be considered as a target likelihood score for any given input. The detailed architecture of the $\mathcal{A}$ network is indicated in the Fig. \ref{fig:A_network}.

\textbf{$\mathcal{R+A}$ Adversarial Training:}
Goodfellow \etal~ \cite{gan} has introduced an efficient way for adversarial learning of two
networks, Generator (G) and Discriminator (D), called GAN. GANs aim to generate samples that follow the real data distribution, through adversarial training of two networks. G learns to map a latent space like Z sampled from a specific distribution \ie,~ $p_z$, to a real data distribution (referred to as $p_t$). D  is trained by maximizing the probability of assigning the correct label to both the actual data and the fake data from G, while G is simultaneously trained to minimize the $\log(1-D(G(Z)))$. In other words, G and D play the following two-players mini-max game:
\begin{equation}
	\begin{split}
		\underset{G}{\min} ~ \underset{D}{\max} ~~ \Big( & \mathbb{E}_{X \sim p_{t}}[\log(D(X))] \\ 
		& + \mathbb{E}_{Z \sim p_{z}}[\log(1-D(G(Z)))]\Big)
	\end{split}
\end{equation}
Similarly, we have jointly and adversarially trained $\mathcal{R}+\mathcal{A}$ network, and on the contrary to the purpose of conventional GANs which learns to generate sample from $p_t$ distribution, $\mathcal{R}$ and $\mathcal{A}$ are trained to generate sample for abnormal class and distinguish abnormal flows from normal ones, respectively. Consequently, the optimum point for stopping the joint training of $\mathcal{R}+\mathcal{A}$ differs from conventional GANs. In beside of this, in our method instead of mapping the latent representation, Z, to a data sample with the distribution $p_{t}$, $\mathcal{R}$ maps:

\begin{equation}
	X \sim p_{t} \longrightarrow X' \nsim p_{t}
\end{equation}

As stated earlier, $p_t$ is the distribution of the target class, \ie,~ normal traffic. Since the PFs from the target class are available for $\mathcal{A}$, thus it knows $p_t$. In this case, $\mathcal{A}$ can explicitly decide whether $\mathcal{R}(X)$ follows $p_t$ or not. Accordingly, $\mathcal{R}+\mathcal{A}$ can be trained by optimizing the following objective:

\begin{equation}
	\begin{split}
		\underset{\mathcal{R}}{\min} ~ \underset{\mathcal{A}}{\max} ~~ \Big( & \mathbb{E}_{X \sim p_{t}}[\log(\mathcal{A}(X))] \\ 
		& + \mathbb{E}_{X \sim p_{t}}[\log(1-\mathcal{A}(\underbrace{\mathcal{R}(X)}_{X'}))]\Big)
	\end{split}
\end{equation}

\textbf{$\mathcal{A}$nomaly Detection:}
In this part, we have explained the classification manner of the proposed method. As previously discussed, $\mathcal{A}$ acts as an anomaly detector, and also derive a benefit from $\mathcal{R}$. 
Eventually, the proposed Anomaly Detector (AD) is formulated merely using $\mathcal{A}$ network as follows:

\begin{equation}
	\begin{split}
		\label{eq:classification}
		\text{AD$(X)$ =} \begin{cases}
			\text{Normal (Target Class)} & \text{if $\mathcal{A}(X)$ $>$ $\alpha$,} \\
			\text{Anomaly (Outlier)} & \text{Otherwise.}
		\end{cases}
	\end{split} 
\end{equation}

Where $\alpha \in (0,1)$ is a threshold value. 

\section{Experimental Results}
In this section, We have evaluated our proposed method on a widely-used dataset known as NSL-KDD\footnote{NSL-KDD is available at https://github.com/defcom17/NSL$\_$KDD} \cite{nsl}. The experimental results along with a thorough comparison are provided in the following subsections.

\subsection{Implementation Details}
The proposed method is implemented using Keras framework and Python ran on the GOOGLE COLAB\footnote{https://colab.research.google.com}. Learning rate is set equal to 0.001 for both networks and $\alpha$ (Equation \ref{eq:classification}) is set equal to 0.5. The detailed structure of $\mathcal{R}$ and $\mathcal{A}$ \footnote{The trained models of both networks, \ie,~ $\mathcal{R}$ and $\mathcal{A}$, are available at https://github.com/Bahram-Mohammadi/End-to-End-Adversarial-Learning-for-Intrusion-Detection-in-Computer-Networks} has been provided in Section \ref{sec:end-to-end}.

\subsection{Evaluation}
As mentioned previously, the assessment is carried out using the NSL-KDD dataset including two subsets, KDDTrain$^+$ and KDDTest$^+$. Since we have proposed a semi-supervised method, only the normal records of KDDTrain$^+$ are involved in the training phase and thus its anomalous records are unused. Furthermore, for the stage of the test, KDDTest$^+$ is used completely. Table \ref{tab:data_distriburion} indicates the data distribution of training and test sets for the evaluation process.

\begin{table}[t]
	\centering
	\caption{Data distribution of training and test sets.}
	\vspace{-2mm}
	\label{tab:data_distriburion}
	\begin{tabular}{|c|c|c|}
		\cline{2-3}
		\multicolumn{1}{c|}{} & \multicolumn{2}{c|}{\textbf{Number of PFs}} \\
		\hline
		\textbf{Class} & \textbf{KDDTrain$^{+}$} & \textbf{KDDTest$^{+}$} \\
		\noalign{\hrule height 1.5pt}
		\textbf{Normal} & 67343 & 9711 \\
		\hline
		\textbf{Anomaly} & --- & 12833 \\
		\hline
	\end{tabular}
\end{table}

\begin{table}[t]
	\centering
	\caption{Binary classification performance Comparison.}
	\vspace{-2mm}
	\label{tab:proposed_method_evaluation}
	\begin{tabular}{c c c c c}
		\hline
		\hline
		
		& $ACC(\%)$ & $PR(\%)$ & $RE(\%)$ & $FS(\%)$ \\ \cline{2-5}
		
		AE \cite{ae_dae} & 88.28 & 91.23 & 87.86 & 89.51 \\
		De-noising AE \cite{ae_dae} & 88.65 & \textbf{96.48} & 83.08 & 89.28 \\
		Ours $\mathcal{A}(X)$ & \textbf{91.39} & 89.94 & \textbf{95.56} & \textbf{92.67} \\
		
		\hline
		\hline
	\end{tabular}
\end{table}

Our method ($\mathcal{A}(X)$) is evaluated using four measures: accuracy (ACC), precision (PR), recall (RE) and f-score (FS). We are able to compare $\mathcal{A}(X)$ with methods using the same test set, \ie,~ KDDTest$^+$, and also report the performance results of their work with same metrics. Table \ref{tab:proposed_method_evaluation} confirms that $\mathcal{A}(X)$ achieves better results compared to another work included two different methods \cite{ae_dae}. Although these methods are semi-supervised, they used anomalous records of KDDTrain$^{+}$ in the validation set. In fact, in this work, the threshold for distinguishing normal traffic from anomalies are determined based on the validation set. But abnormal samples have not been involved in the training phase of $\mathcal{A}(X)$ at all. Nonetheless, we have gained better results in terms of ACC, RE, and FS.

\begin{table}[t]
	\centering
	
	\caption{Binary Classification Accuracy Comparison. There are two kinds of Supervision for Methods in This table, Supervised (S) and Semi-Supervied (SS).}
	\vspace{-3mm}
	\label{tab:accuracy-comparison}
	\begin{tabular}{lccc}
		\hline
		
		Method & \multicolumn{2}{c}{Supervision} &  ACC (\%) \\
		& S & SS & \\ \hline
		
		RNN-IDS \cite{rnn_ids} & \checkmark & & 83.28 \\	
		
		DCNN \cite{lstm_dcnn} & \checkmark & & 85.00 \\
		
		
		AE \cite{ae_dae} &  & \checkmark & 88.28 \\
		
		
		Sparse AE and MLP \cite{sae_mlp} & \checkmark & & 88.39 \\
		
		
		Random Tree \cite{randomTree} & \checkmark & & 88.46 \\
		
		
		De-noising AE \cite{ae_dae} & & \checkmark & 88.65 \\
		
		
		LSTM \cite{lstm_dcnn} & \checkmark & & 89.00 \\
		
		
		Random Tree and NBTree \cite{randomTree} & \checkmark & & 89.24 \\
		
		
		
		
		Ours $ \mathcal{A}(X) $ & & \checkmark & \textbf{91.39} \\

		\hline
	\end{tabular}
\end{table}

The selected performance indicator for providing a thorough comparison with the other state-of-the-art methods is $ACC$ showing the correct classification rate of all classes. Regarding Table \ref{tab:accuracy-comparison}, $\mathcal{A}(X)$ represents a significant improvement in terms of accuracy. In our work, simulated abnormal flows are generated with some deviation from the target class irrespective of real anomalies. In fact, our decision making process is not biased towards intrusions which are available in the training set. Hence, $\mathcal{A}(X)$ includes generalization property helping us to obtain better result even compared to supervised methods. It is worth mentioning that, the process of detecting each incoming network PF is done just in 45$\mu s$ on average. Therefore, $\mathcal{A}(X)$ is capable of properly working in real time.
\section{Discussion}
\textbf{Mode collapse:}
GANs face an issue arising when the generator learns only a portion of real-data distribution and then outputs samples from a single mode, (\ie,~ the other modes are ignored). This problem is known as mode collapse \cite{modeCollapse}. Mode collapse is no longer exists in our case as $\mathcal{R}(X)$ directly sees all possible flows of the target class data and implicitly learns the manifold spanned by the target data distribution.

\textbf{Unseen class generating:}
Our proposed method is semi-supervised, thus we need to somehow obtain the anomalous PFs from the normal traffic. The simulated anomalous traffic are generated by $\mathcal{R}$ network to play the role of real anomalies. In fact, we have approximately generated the unseen class data using the normal traffic to solve the problem of DNNs with the absence of one class data.

\textbf{Generalization:}
If the training process is done irrespective of the anomalies in the training set, the proposed method is able to decide about the type of incoming PFs while it is not biased towards the already known intrusions. In this case, the proposed method can provide generalization leading to better detection rate.
\section{conclusion}
In this paper, we have proposed a novel semi-supervised and end-to-end deep learning method for anomaly detection in IDS. Specifically, our model is composed of two modules, $\mathcal{R}$ and $\mathcal{A}$. These networks are trained competitively in an adversarial manner. After training phase, $\mathcal{R}$ can simulate anomalies in a way that do not perfectly match with the normal traffic, while $\mathcal{A}$ can distinguish normal PFs from abnormal traffic. The proposed method has been evaluated on NSL-KDD dataset and the results demonstrates our method has better performance compared to the other state-of-the-arts.

\ifCLASSOPTIONcaptionsoff
\newpage
\fi
\nocite{*}
\bibliographystyle{IEEEtran}
{\small
	\bibliography{References} 
}



%



\end{document}